\documentclass[graybox,a4paper]{svmult}
\usepackage{geometry}
\usepackage{url}

\usepackage[resetlabels]{multibib} 

\usepackage{tikz} 

\usepackage{amsfonts}   
\usepackage{amsmath}    
\usepackage{siunitx}    
\usepackage{newtxtext,newtxmath}  
\usepackage{helvet}         
\usepackage{courier}        
\usepackage{pifont}      

\usepackage{stmaryrd}   
\usepackage{trimclip}   
\usepackage{layouts}

\usepackage{booktabs}
\usepackage{makecell}  
\usepackage[flushleft]{threeparttable}  
\usepackage{multirow}

\usepackage[inline]{enumitem} 
\usepackage{xcolor}    
\usepackage{textcomp}  
\usepackage{gensymb}   
\usepackage{graphicx}  
\usepackage[font=footnotesize, skip=5pt, hypcap=false]{caption}
\usepackage[caption=false,font=footnotesize]{subfig}
\usepackage[pdfencoding=auto, hidelinks]{hyperref} 

\newcites{SM}{Supplementary Material References}

\makeatletter
\DeclareRobustCommand{\shortto}{%
  \mathrel{\mathpalette\short@to\relax}%
}
\newcommand{\short@to}[2]{%
  \mkern2mu
  \clipbox{{.25\width} 0 0 0}{$\m@th#1\vphantom{+}{\shortrightarrow}$}%
  }
\makeatother

\newcommand{\refeqn}[1]{Eq.~\ref{#1}}
\newcommand{\reffig}[1]{Fig.~\ref{#1}}
\newcommand{\refsec}[1]{Sec.~\ref{#1}}

\newcommand{\reftab}[1]{Table~\ref{#1}}

\usepackage{titlesec}
\titlespacing*{\section}{0pt}{1.2\baselineskip}{\baselineskip}
\titlespacing*{\subsection}{0pt}{1.2\baselineskip}{\baselineskip}

\makeindex             

\begin{document}

\title*{Continual SLAM: Beyond Lifelong Simultaneous Localization and Mapping through \texorpdfstring{\\}{} Continual Learning}
\titlerunning{Continual SLAM -- Accepted at ISRR 2022, preprint version with suppl. material}  
\author{
Niclas Vödisch\inst{1}, 
Daniele Cattaneo\inst{1}, 
Wolfram Burgard\inst{2}, and 
Abhinav Valada\inst{1}
}
\institute{
\inst{1}Department of Computer Science, University of Freiburg, Germany,\\
\inst{2}Department of Engineering, University of Technology Nuremberg, Germany \\
This work was funded by the European Union’s Horizon 2020 research and innovation program under grant agreement No 871449-OpenDR. \\
This is a preprint of the following chapter: N. Vödisch, D. Cattaneo, W. Burgard \& A. Valada, Continual SLAM: Beyond Lifelong Simultaneous Localization and Mapping Through Continual Learning, published in Robotics Research, edited by A. Billard, T. Asfour \& O. Khatib, 2023, Springer reproduced with permission of Springer. The final authenticated version is available online at: \url{https://doi.org/10.1007/978-3-031-25555-7_3}.
}

\maketitle              


\abstract{
    Robots operating in the open world encounter various different environments that can substantially differ from each other. This domain gap also poses a challenge for Simultaneous Localization and Mapping (SLAM) being one of the fundamental tasks for navigation. In particular, learning-based SLAM methods are known to generalize poorly to unseen environments hindering their general adoption. In this work, we introduce the novel task of continual SLAM extending the concept of lifelong SLAM from a single dynamically changing environment to sequential deployments in several drastically differing environments. To address this task, we propose CL-SLAM leveraging a dual-network architecture to both adapt to new environments and retain knowledge with respect to previously visited environments. We compare CL-SLAM to learning-based as well as classical SLAM methods and show the advantages of leveraging online data. We extensively evaluate CL-SLAM on three different datasets and demonstrate that it outperforms several baselines inspired by existing continual learning-based visual odometry methods. We make the code of our work publicly available at \url{http://continual-slam.cs.uni-freiburg.de}.

}


\section{Introduction}
\label{sec:introduction}

\begin{figure}
    \centering
    \includegraphics[width=0.9\linewidth]{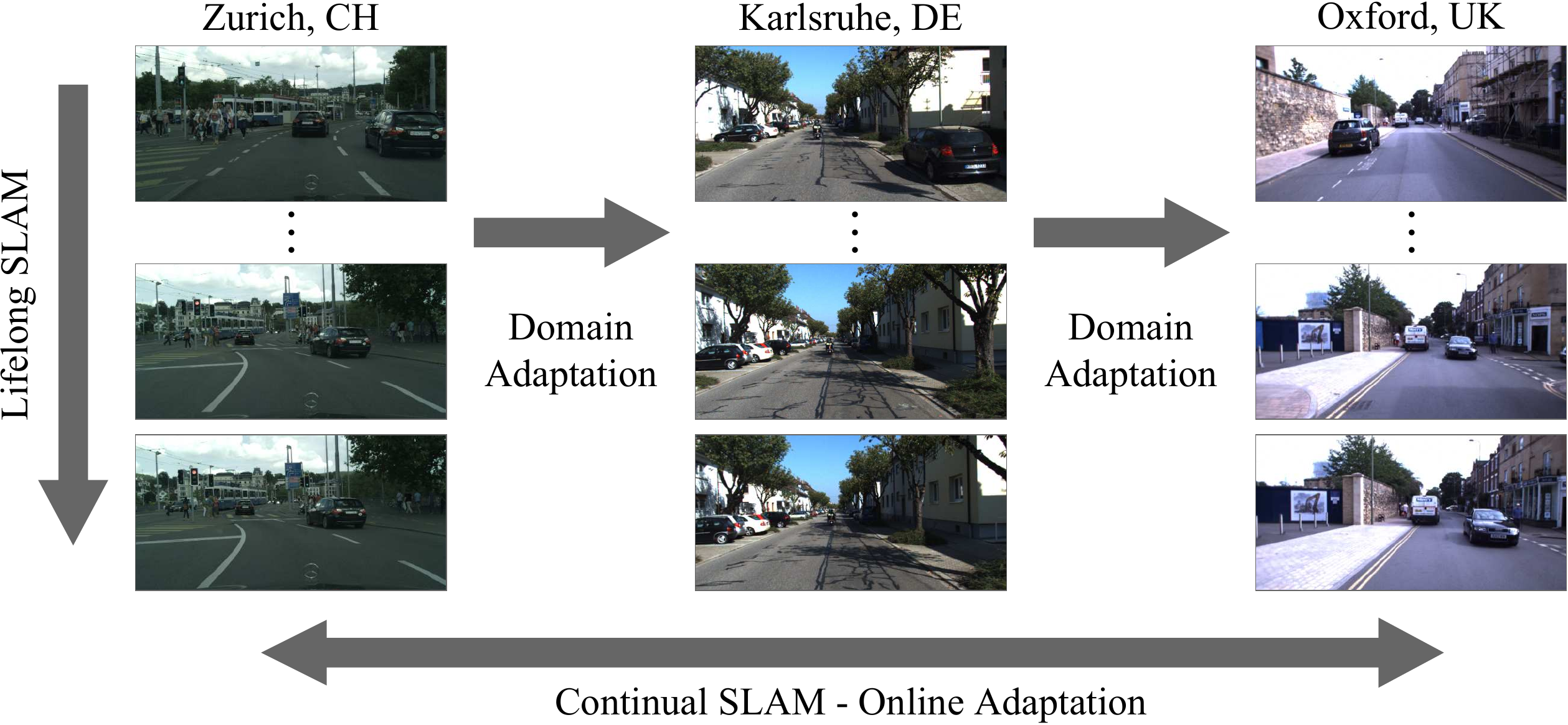}
    \caption{While lifelong SLAM considers the long-term operation of a robot in a single dynamically changing environment, domain adaptation techniques aim toward transferring knowledge gained in one environment to another environment. The newly defined task of continual SLAM extends both settings by requiring omnidirectional adaptation involving multiple environments. Agents have to both quickly adapt to new environments and effectively recall knowledge from previously visited environments.}
    \label{fig:cover}
    \vspace{-0.5cm}
\end{figure}

An essential task for an autonomous robot deployed in the open world without prior knowledge about its environment is to perform Simultaneous Localization and Mapping (SLAM) to facilitate planning and navigation~\cite{hurtado2021learning, mittal2019vision}. To address this task, various SLAM algorithms based on different sensors have been proposed, including classical methods~\cite{murartal2015orbslam} and learning-based approaches~\cite{cattaneo2021lcdnet,li2021deepslam}. Classical methods typically rely on handcrafted low-level features that tend to fail under challenging conditions, e.g., textureless regions. Deep learning-based approaches mitigate such problems due to their ability to learn high-level features. However, they lack the ability to generalize to out-of-distribution data, with respect to the training set. For visual SLAM, such out-of-distribution data can correspond to images sourced from cities in different countries or under substantially different conditions. In the following, we use the term \textit{environment} to refer to a bounded geographical area. While different environments can share the same fundamental structure, e.g., urban areas, their specific characteristics prevent the seamless transfer of learned features, resulting in a domain gap between cities~\cite{bevsic2021unsupervised}.

In the context of this work, lifelong SLAM~\cite{thrun1995is} considers the long-term operation of a robot in a single environment (see \reffig{fig:cover}). Although this environment can be subject to temporal changes, the robot is constrained to stay within the area borders~\cite{kretzschmar2010lifelong}, e.g., to obtain continuous map updates~\cite{kurz2021geometry} within a city. Recent works attempt to relax this assumption by leveraging domain adaptation techniques for deep neural networks, including both regularization~\cite{zhang2020online} and online adaptation of the employed model~\cite{li2020self, li2021generalizing, luo2019real}. While a naive solution for adapting to a new environment is to source additional data, this is not feasible when the goal is to ensure the uninterrupted operation of the robot. Moreover, changes within an environment can be sudden, e.g., rapid weather changes, and data collection and annotation often come at a high cost. Therefore, adaptation methods should be trainable in an unsupervised or self-supervised manner without the need for ground truth data. As illustrated in \reffig{fig:cover}, the setting addressed in domain adaptation only considers unidirectional knowledge transfer from a single known to a single unknown environment~\cite{bevsic2021unsupervised} and thus does not represent the open world, where the number of new environments that a robot can encounter is infinite and previously seen environments can be revisited. To address this gap, we take the next step by considering more complex sequences of environments and formulate the novel task of continual SLAM that leverages insights from both continual learning (CL) and lifelong SLAM.
We propose a dual-network architecture called CL-SLAM to balance adaptation to new environments and memory retention of preceding environments. To assess its efficacy, we define two metrics, adaptation quality and retention quality, and compare CL-SLAM to several baselines inspired by existing CL-based VO methods involving three different environments.
We make the code of this work publicly available at \url{http://continual-slam.cs.uni-freiburg.de}.
The supplementary material can be found at \url{https://arxiv.org/abs/2203.01578}.

\section{Related Work}


\noindent\textit{Visual Odometry / SLAM:}
Visual odometry (VO) and vision-based SLAM estimate camera motion from a video. Allowing for self-supervised training, monocular VO can be tackled jointly with depth estimation based on photometric consistency. SfMLearner~\cite{zhou2017unsupervised} uses an end-to-end approach consisting of two networks to predict depth from a single image and camera motion from two consecutive images. The networks are trained in parallel by synthesizing novel views of the target image. Monodepth2~\cite{godard2019digging} extends the loss function to account for occluded and static pixels. Other works such as DF-VO~\cite{zhan2020visual} eliminate the need for a pose network by leveraging feature matching based on optical flow. While these methods show superior performance~\cite{li2021generalizing}, computing a gradient of the predicted pose with respect to the input image is not possible using classic point matching algorithms. To reduce drift, DeepSLAM~\cite{li2021deepslam} combines unsupervised learning-based VO with a pose graph backend taking global loop closures into account. In this work, we use a trainable pose network with velocity supervision~\cite{guizilini20203d} to resolve scale ambiguity. Similar to DeepSLAM, we detect loop closures and perform graph optimization.


{\parskip=3pt
\noindent\textit{Continual Learning:}
Traditionally, a learning-based model is trained for a specific task on a dedicated training set and then evaluated on a hold-out test set sampled from the same distribution. However, in many real-world applications, the data distributions can differ or even change over time. Additionally, the initial task objective might be altered. Continual learning (CL) and lifelong learning~\cite{thrun1995is} address this problem by defining a paradigm where a model is required to continuously readjust to new tasks and/or data distributions without sacrificing the capability to solve previously learned tasks, thus avoiding catastrophic forgetting.
Most CL approaches employ one of three strategies.
First, experience replay includes rehearsal and generative replay. Rehearsal refers to reusing data samples of previous tasks during adaptation to new tasks, e.g., the replay buffer in CoMoDA~\cite{kuznietsov2021comoda}. Minimizing the required memory size, the most representative samples should be carefully chosen or replaced by more abstract representations~\cite{gopalakrishnan2022knowledge}. Similarly, generative replay constructs artificial samples by training generative models~\cite{shin2017continual}.
Second, regularization~\cite{li2018learning} prevents a CL algorithm from overfitting to the new tasks to mitigate forgetting, e.g., knowledge distillation.
Third, architectural methods~\cite{kemker2018fearnet} preserve knowledge by adding, duplicating, freezing, or storing parts of the internal model parameters. They further include dual architectures that are inspired by mammalian brains~\cite{mcclelland1995why}, where one model learns the novel task and a second model memorizes previous experience.
In this work, we combine architectural and replay strategies by leveraging a dual-network architecture with online adaptation incorporating data rehearsal.
}


{\parskip=3pt
\noindent\textit{Online Adaptation for Visual Odometry and Depth Estimation:}
Recently, Luo~\textit{et al.}~\cite{luo2019real} employed a subtask of CL for self-supervised VO and depth estimation, opening a new avenue of research. Online adaptation enables these methods to enhance the trajectory and depth prediction on a test set sourced from a different data distribution than the originally used training set. Both Zhang~\textit{et~al.}~\cite{zhang2020online} and CoMoDA~\cite{kuznietsov2021comoda} primarily target the depth estimation task. While Zhang~\textit{et~al.} propose to learn an adapter to map the distribution of the online data to the one of the training data, CoMoDA updates the internal parameters of the depth and pose networks based on online data and a replay buffer.
The work in spirit most similar to ours is done by Li~\textit{et~al.}~\cite{li2020self}. They propose to substitute the standard convolutional layers in the depth and pose networks with convolutional LSTM variants. Then, the model parameters are continuously updated using only the online data. In subsequent work, Li~\textit{et~al.}~\cite{li2021generalizing} replace the learnable pose network by point matching from optical flow. Note that all existing works purely focus on one-step adaptation, i.e., transferring knowledge gained in one environment to a single new environment.
In this paper, we introduce continual SLAM to take the next step by considering more complex deployment scenarios comprising more than two environments and further alternating between them.
}

\section{Continual SLAM}


\noindent\textbf{Problem Setting:} 
\label{ssec:problem_setting}
Deploying a SLAM system in the open world substantially differs from an experimental setting, in which parameter initialization and system deployment are often performed in the same environment.
To overcome this gap, we propose a new task called \textit{Continual SLAM}, illustrated in \reffig{fig:cover}, where the robot is deployed on a sequence of diverse scenes from different environments. 

Ideally, a method addressing the continual SLAM problem should be able to achieve the following goals:
\begin{enumerate*}[label={\arabic*)}]
\item quickly adapt to unseen environments while deployment,
\item leverage knowledge from previously seen environments to speed up the adaptation, and 
\item effectively memorize knowledge from previously seen environments to minimize the required adaptation when revisiting them, while mitigating overfitting to any of the environments.
\end{enumerate*}
Formally, continual SLAM can be defined as a potentially infinite sequence of scenes $\mathcal{S} = ( s_1 \to s_2 \to \dots )$ from a set of different environments $s_i \in \{ E_a, E_b, \dots \}$, where $s$ denotes a scene and $E$ denotes an environment.
In particular, $\mathcal{S}$ can contain multiple scenes from the same environment and the scenes in $\mathcal{S}$ can occur in any possible fixed order. A continual SLAM algorithm $\mathcal{A}$ can be defined as\looseness=-1
\vspace*{-.2cm}
\begin{equation}
    \mathcal{A}: \hspace{0.5em} < \theta_{i-1}, (s_1, \dots, s_i ) > \hspace{0.5em} \mapsto \hspace{0.5em} < \theta_i >,
\end{equation}

\vspace*{-.2cm} \noindent where $( s_1, \dots, s_i )$ refers to the seen scenes in the specified order and $\theta_i$ denotes the corresponding state of the learnable parameters of the algorithm. During deployment, the algorithm $\mathcal{A}$ has to update $\theta_{i-1}$ based on the newly encountered scene $s_i$. For instance, given two environments $E_a = \{s_a^1, s_a^2\}$ and $E_b = \{s_b^1\}$, which comprise two and one scenes, respectively, examples of feasible sequences are
\vspace*{-.2cm}
\begin{equation}
\begin{aligned}
    \mathcal{S}_1 = (s_a^1 \to s_b^1 \to s_a^2), 
    \hspace{12pt}
    \mathcal{S}_2 = (s_a^2 \to s_a^1 \to s_b^1),
    \hspace{12pt}
    \mathcal{S}_3 = (s_b^1 \to s_a^2 \to s_a^1),
\end{aligned}
\end{equation}

\vspace*{-.2cm} \noindent where the scene subscripts denote the corresponding environment and the superscripts refer to the scene ID in this environment.
As described in \refsec{sec:introduction}, the task of continual SLAM is substantially different from lifelong SLAM or unidirectional domain adaptation as previously addressed by Luo~\textit{et~al.}~\cite{luo2019real} and Li~\textit{et~al.}~\cite{li2020self, li2021generalizing}.

To conclude, we identify the following main challenges:
1) large number of different environments, 2) huge number of chained scenes, 3) scenes can occur in any possible order, and 4) environments can contain multiple scenes.
Therefore, following the spirit of continual learning (CL), a continual SLAM algorithm has to balance between short-term adaptation to the current scene and long-term knowledge retention. This trade-off is also commonly referred to as avoiding catastrophic forgetting with respect to previous tasks without sacrificing performance on the new task at hand.

{\parskip=3pt
\noindent\textbf{Performance Metrics:} 
\label{ssec:metrics}
To address the aforementioned challenges, we propose two novel metrics, namely adaptation quality (AQ), which measures the short-term adaptation capability when being deployed in a new environment, and retention quality (RQ), which captures the long-term memory retention when revisiting a previously encountered environment. In principle, these metrics can be applied to any given base metric $M_d$ that can be mapped to the interval $[0, 1]$, where 0 and 1 are the lowest and highest performances, respectively. The subscript $d$ denotes the given sequence, where the error is computed on the final scene.}


{\parskip=3pt
\noindent\textit{Base Metrics:}
For continual SLAM, we leverage the translation error $t_{\mathit{err}}$ (in \%) and the rotation error $r_{\mathit{err}}$ (in \degree/m), proposed by Geiger~\textit{et al.}~\cite{geiger2012are}, that evaluate the error as a function of the trajectory length. To obtain scores in the interval $[0, 1]$, we apply the following remapping:
\vspace*{-.2cm}
\begin{equation}
\begin{aligned}
    \widehat t_{\mathit{err}} &= \max \left( 0, 1 - \frac{t_{\mathit{err}}}{100} \right),
    \hspace{12pt}
    \widehat r_{\mathit{err}} &= 1 - \frac{r_{\mathit{err}}}{180},
\end{aligned}
\label{eqn:err_remapping}
\end{equation}

\vspace*{-.2cm} \noindent where we clamp $\widehat t_{\mathit{err}}$ to 0 for $t_{\mathit{err}} > 100\%$. The resulting $\widehat t_{\mathit{err}}$ and $\widehat r_{\mathit{err}}$ are then used as the base metric $M$ to compute AQ\textsubscript{trans} / RQ\textsubscript{trans} and AQ\textsubscript{rot} / RQ\textsubscript{rot}, respectively.
}


{\parskip=3pt
\noindent\textit{Adaptation Quality:}
The adaptation quality (AQ) measures the ability of a method to effectively adapt to a new environment based on experiences from previously seen environments. It is inspired by the concept of forward transfer (FWT)~\cite{lopez2017gradient} in traditional CL, which describes how learning a current task influences the performance of a future task. Particularly, positive FWT enables zero-shot learning, i.e., performing well on a future task without explicit training on it. On the other hand, negative FWT refers to sacrificing performance on a future task by learning the current task. In our context, a task refers to performing SLAM in a given environment. Consequently, the AQ is intended to report how well a continual SLAM algorithm is able to minimize negative FWT, e.g., by performing online adaptation.
}

To illustrate the AQ, we consider the simplified example of a set of two environments $\{ E_a, E_b \}$ consisting of different numbers of scenes. We further assume that the algorithm has been initialized in a separate environment $E_p$. Since the AQ focuses on the cross-environment adaptation, we sample one random scene from each environment $s_a \in E_a$ and $s_b \in E_b$ and hold them fixed. Now, we construct the set of all possible deployment sequences $\mathcal{D} = \left\{ (s_p \to s_a), (s_p \to s_b), (s_p \to s_a \to s_b), (s_p \to s_b \to s_a) \right\}$, where $s_p \in E_p$ is the data used for initialization.
The AQ is then defined as:
\vspace*{-.2cm}
\begin{equation}
    \text{AQ} = \frac{1}{|\mathcal{D}|} \sum_{d \in \mathcal{D}} M_d.
\end{equation}


{\noindent\textit{Retention Quality:}
To further account for the opposing challenge of the continual SLAM setting, we propose the retention quality (RQ) metric. It measures the ability of an algorithm to preserve long-term knowledge when being redeployed in a previously encountered environment. It is inspired by the concept of backward transfer (BWT)~\cite{lopez2017gradient} in CL settings, which describes how learning a current task influences the performance on a previously learned task. While positive BWT refers to improving the performance on prior tasks, negative BWT indicates a decrease in the performance of the preceding task. The extreme case of a large negative BWT is often referred to as catastrophic forgetting.
Different from classical BWT, we further allow renewed online adaptation when revisiting a previously seen environment, i.e., performing a previous task, as such a setting is more sensible for a robotic setup. It further avoids the necessity to differentiate between new and already seen environments, which would require the concept of environment classification in the open world.
}

To illustrate the RQ, we consider a set of two environments $\{ E_a, E_b \}$ consisting of different numbers of scenes. We further assume that the algorithm has been initialized on data $s_p$ of a separate environment $E_p$. We sample two random scenes from each environment, i.e., $s_a^1$, $s_a^2$, $s_b^1$, and $s_b^2$.
To evaluate the RQ, we need to construct a deployment sequence $S$ that consists of alternating scenes from the two considered environments. In this example, we consider the following fixed sequence:
\vspace*{-.2cm}
\begin{equation}
\begin{aligned}
    S = ( s_p \to s_a^1 \to s_b^1 \to s_a^2 \to s_b^2 ).
\end{aligned}
\label{eqn:rq_seq}
\end{equation}

\vspace*{-.2cm} \noindent
We then consider all the subsequences $\mathcal{D}$ of $S$ in which the last scene comes from an environment already visited prior to a deployment in a scene of a different environment. In this example,  $\mathcal{D} = \big\{ (s_p \to s_a^1 \to s_b^1 \to s_a^2),$ $(s_p \to s_a^1 \to s_b^1 \to s_a^2 \to s_b^2) \big\}$.

The RQ is then defined as the sum over all differences of the base metric in a known environment before and after deployment in a new environment, divided by the size of $\mathcal{D}$. For instance, given the sequence in \refeqn{eqn:rq_seq}:
\vspace*{-.2cm}
\begin{equation}
\begin{aligned}
    \text{RQ} &= \frac{1}{2} \left( M_{s_p \shortto s_a^1 \shortto s_b^1 \shortto s_a^2} - M_{s_p \shortto s_a^1 \shortto s_a^2} + M_{s_p \shortto s_a^1 \shortto s_b^1 \shortto s_a^2 \shortto s_b^2} - M_{s_p \shortto s_a^1 \shortto s_b^1 \shortto s_b^2} \right).
\end{aligned}
\label{eqn:rq}
\end{equation}

\section{Technical Approach}

\begin{figure*}[t]
    \centering
    \includegraphics[width=\linewidth]{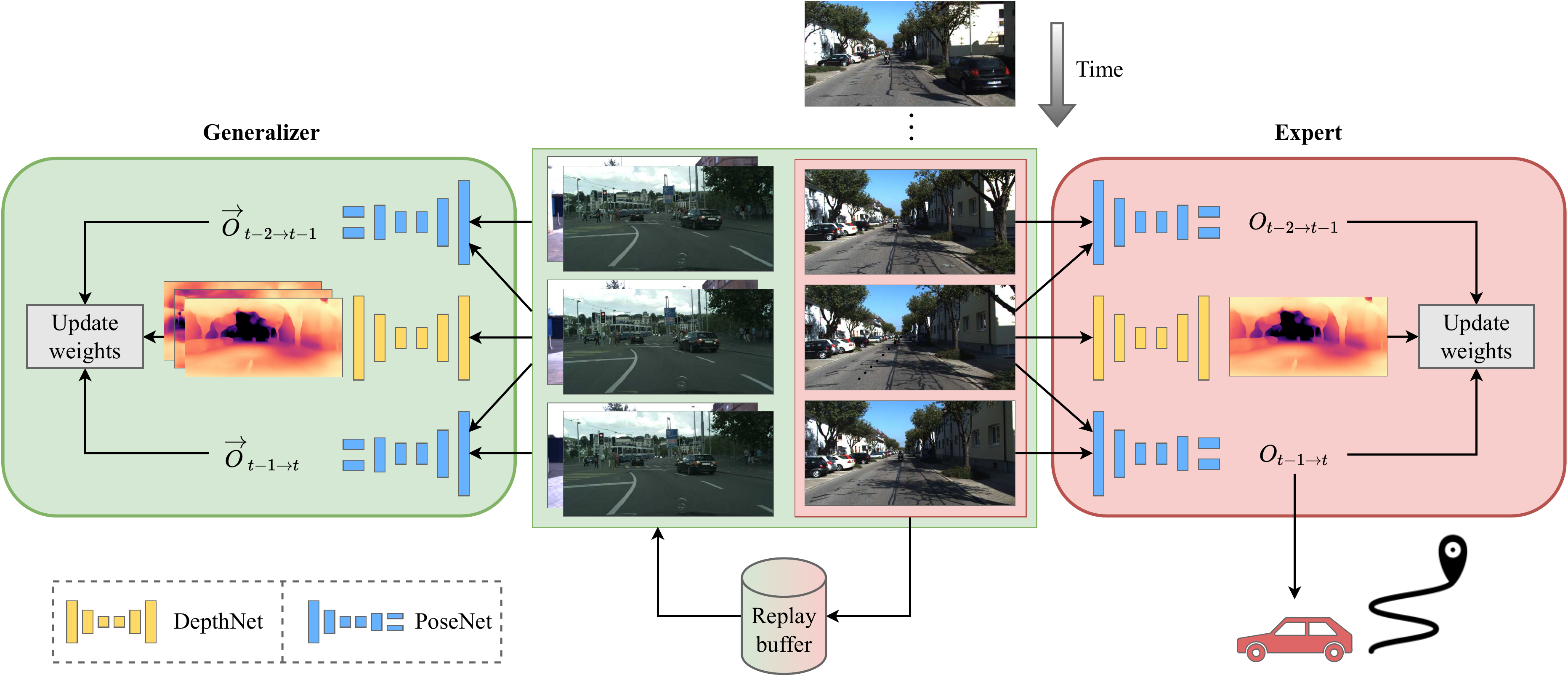}
    \vspace*{-.5cm}
    \caption{Online adaptation scheme of our proposed CL-SLAM that is constructed as a dual-network architecture including a generalizer (left) and an expert (right). While the expert focuses on the short-term adaptation to the current scene, the generalizer avoids catastrophic forgetting by employing a replay buffer comprising samples from the past and the present. Note that both subnetworks contain a single PoseNet, shown twice to reflect pose estimation at different steps. The predicted odometry $O_{t-1 \shortto t}$ is sent to the SLAM framework as shown in \reffig{fig:overview_slam}.}
    \label{fig:overview}
    \vspace*{-.5cm}
\end{figure*}
\begin{figure*}[t]
    \centering
    \includegraphics[width=\linewidth]{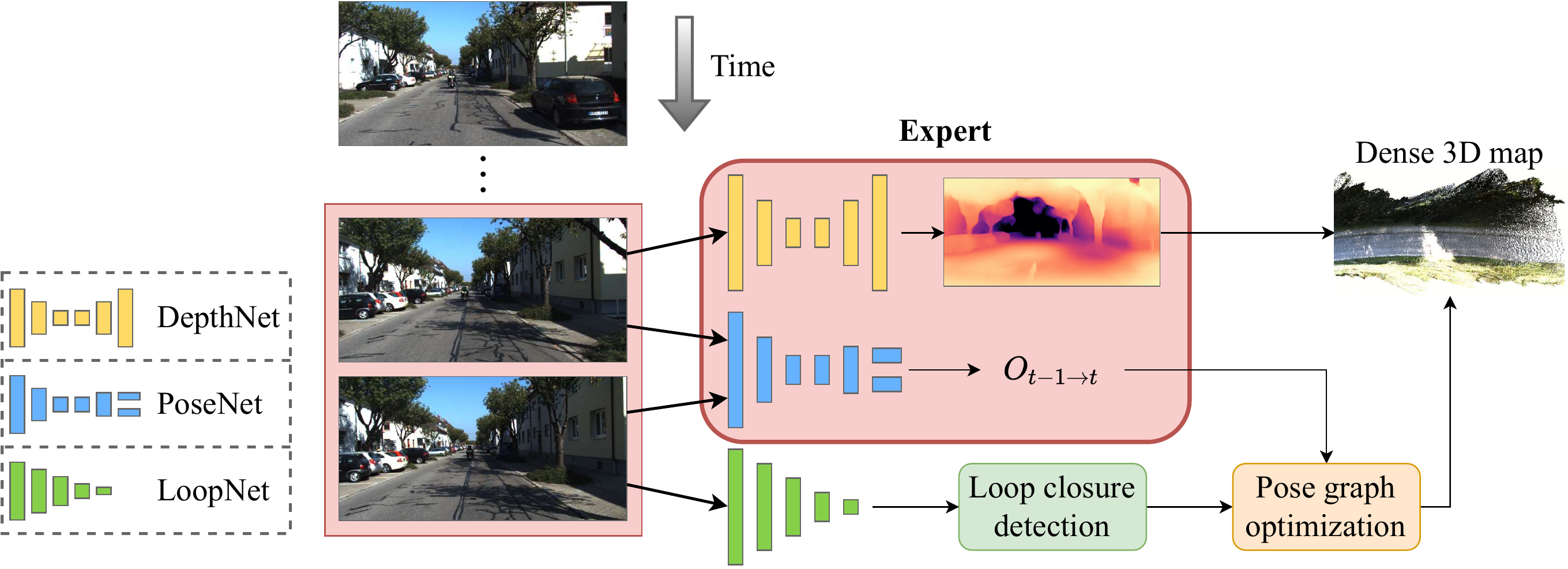}
    \vspace*{-.4cm}
    \caption{Full SLAM framework of our proposed CL-SLAM. Global loop closures are detected with a pre-trained CNN. Visual odometry estimates between both consecutive frames and loop closure frames are generated by the PoseNet and added to a pose graph, which is optimized upon the detection of a new loop closure. Finally, a dense 3D map can be created using the predicted depth and the optimized path.}
    \label{fig:overview_slam}
    \vspace*{-.5cm}
\end{figure*}

\noindent\textbf{Framework Overview:} 
The core of CL-SLAM is the dual-network architecture of the visual odometry (VO) model that consists of an \textit{expert} that produces myopic online odometry estimates and a \textit{generalizer} that focuses on the long-term learning across environments (see \reffig{fig:overview}). We train both networks in a self-supervised manner where the weights of the expert are updated only based on online data, whereas the weights of the generalizer are updated based on a combination of data from both the online stream and a replay buffer.
We use the VO estimates of the expert to construct a pose graph (see \reffig{fig:overview_slam}). To reduce drift, we detect global loop closures and add them to the graph, which is then optimized. Finally, we can create a dense 3D map using the depth predicted by the expert and the optimized path.

\label{ssec:visual_odometry}
{\parskip=3pt
\noindent\textbf{Visual Odometry:}
We generate VO estimates following the commonly used approach of using a trainable pose network~\cite{besic2022dynamic, godard2019digging, guizilini20203d, li2021deepslam} for self-supervised depth estimation with a stream of monocular images. The basic idea behind this approach is to synthesize a novel view of an input image using image warping as reviewed in the supplementary material.
}

In this work, we use Monodepth2~\cite{godard2019digging} to jointly predict the depth map of an image and the camera motion from the previous timestep to the current. To recover metric scaling of both depth and the odometry estimates, we adapt the original loss function with a velocity supervision term as proposed by Guizilini \textit{et al.}~\cite{guizilini20203d}. As scalar velocity measurements are commonly available in robotic systems, e.g., by wheel odometry, this does not pose an additional burden.
Our total loss is composed of the photometric reprojection loss $\mathcal{L}_{pr}$, the image smoothness loss $\mathcal{L}_{sm}$, and the velocity supervision loss $\mathcal{L}_{vel}$:
\vspace*{-.2cm}
\begin{equation}
    \label{eqn:total_loss}
    \mathcal{L} = \mathcal{L}_{pr} + \gamma \mathcal{L}_{sm} + \lambda \mathcal{L}_{vel}.
\end{equation}

\vspace*{-.2cm}
Following the common methodology, we compute the loss based on an image triplet $\{\mathbf{I_{t-2}}, \mathbf{I_{t-1}}, \mathbf{I_{t}}\}$ using depth and odometry predictions $\mathbf{D_{t-1}}$, $\mathbf{O_{t-2 \shortto t-1}}$, and  $\mathbf{O_{t-1 \shortto t}}$. We provide more details on the individual losses in the supplementary material.


\label{ssec:loop_closure_graph_optimization}
{\parskip=3pt
\noindent\textbf{Loop Closure Detection and Pose Graph Optimization:}
In order to reduce drift over time, we include global loop closure detection and pose graph optimization (see \reffig{fig:overview_slam}). We perform place recognition using a pre-trained and frozen CNN, referred to as LoopNet.
In particular, we map every frame to a feature vector using MobileNetV3 small~\cite{howard2019searching}, trained on ImageNet, and store them in a dedicated memory. Then, we compute the cosine similarity of the current feature map with all preceding feature maps:}
\vspace*{-.2cm}
\begin{equation}
    \text{sim}_{\cos} = \cos (f_{\mathit{current}}, f_{\mathit{previous}}).
\end{equation}

\vspace*{-.2cm}
If $\text{sim}_{\cos}$ is above a given threshold, we use the PoseNet to compute the transformation between the corresponding images.
During deployment, we continuously build a pose graph~\cite{kummerle2011g2o} consisting of both local and global connections, i.e., consecutive VO estimates and loop closures. Whenever a new loop closure is detected, the pose graph is optimized.

{\parskip=3pt
\noindent\textbf{Online Adaptation:} 
\label{ssec:online_adaptation}
In this section, we describe the dual-network architecture of the VO predictor in CL-SLAM that effectively addresses the trade-off between short-term adaptation and long-term memory retention, a problem also known as catastrophic forgetting. Subsequently, we detail the training scheme including the utilized replay buffer.}


{\parskip=3pt
\noindent\textit{Architecture:}
The dual-network architecture consists of two instances of both the DepthNet and the PoseNet. In the following, we refer to these instances as \textit{expert} and \textit{generalizer}. We build upon the architecture of Monodepth2~\cite{godard2019digging}. The DepthNet has an encoder-decoder topology, comprising a ResNet-18~\cite{he2016deep} encoder and a CNN-based decoder with skip connections, and predicts disparity values for each pixel in the input image. The PoseNet consists of a similar structure using a separate ResNet-18 encoder followed by additional convolutional layers to generate the final output that represents translation and rotation between two input images. Further implementation details are provided in \refsec{ssec:implementation_details}.}


{\parskip=3pt
\noindent\textit{Training Scheme:}
Before deployment, i.e., performing continual adaptation, we pre-train the DepthNet and the PoseNet using the standard self-supervised training procedure based on the loss functions described in \refsec{ssec:visual_odometry}. When deployed in a new environment, we continuously update the weights of both the expert and the generalizer in an online manner, following a similar scheme as Kuznietsov~\textit{et~al.}~\cite{kuznietsov2021comoda}:}
\begin{enumerate}[label={(\arabic*)}]
    \item Create an image triplet composed of the latest frame $\mathbf{I_t}$ and the two previous frames $\mathbf{I_{t-1}}$ and $\mathbf{I_{t-2}}$. Similarly, batch the corresponding velocity measurements.
    \item Estimate the camera motion between both pairs of subsequent images, i.e., $\mathbf{O_{t-2 \shortto t-1}}$ and $\mathbf{O_{t-1 \shortto t}}$ with the PoseNet.
    \item Generate the depth estimate $\mathbf{D_{t-1}}$ of the previous image with the DepthNet.
    \item Compute the loss according to \refeqn{eqn:total_loss} and use backpropagation to update the weights of the DepthNet and PoseNet.
    \item Loop over steps (2) to (4) for $c$ iterations.
    \item Repeat the previous steps for the next image triplet.
\end{enumerate}

Upon deployment, both the expert and the generalizer are initialized with the same set of parameter weights, initially obtained from pre-training and later replaced by the memory of the generalizer. As illustrated in \reffig{fig:overview}, the weights of the expert are updated according to the aforementioned algorithm. Additionally, every new frame from the online image stream is added to a replay buffer along with the corresponding velocity reading. Using only the online images, the expert will quickly adapt to the current environment. This behavior can be described as a desired form of overfitting for a myopic increase in performance.
On the other hand, the generalizer acts as the long-term memory of CL-SLAM circumventing the problem of catastrophic forgetting in continual learning settings. Here, in step (1), we augment the online data by adding image triplets from the replay buffer to rehearse experiences made in the past, as depicted in \reffig{fig:overview}. After deployment, the weights of the stored parameters used for initialization are replaced by the weights of the generalizer, thus preserving the continuous learning process of CL-SLAM. The weights of the expert are then discarded.

\section{Experimental Evaluation}


\label{ssec:implementation_details}

\noindent\textbf{Implementation Details:}
We adopt the Monodepth2~\cite{godard2019digging} architecture using separate ResNet-18~\cite{he2016deep} encoders for our DepthNet and PoseNet. We implement CL-SLAM in PyTorch~\cite{paszke2019pytorch} and train on a single NVIDIA TITAN X GPU. We pre-train both sub-networks in a self-supervised manner on the Cityscapes dataset~\cite{cordts2016the} for 25 epochs with a batch size of 18. We employ the Adam optimizer with $\beta_1=0.9$, $\beta_2=0.999$ and an initial learning rate of $10^{-4}$, which is reduced to $10^{-5}$ after 15 epochs. Further, we resize all images during both pre-training and adaptation to $192 \times 640$ pixels. Additionally, during the pre-training phase, we mask all potentially dynamic objects using bounding boxes generated by YOLOv5m~\cite{jocher2022yolo}, which was trained on the COCO dataset. We observe that on Cityscapes this procedure yields a smaller validation loss than without masking. We set the minimum predictable depth to $\SI{0.1}{\meter}$ without specifying an upper bound. To balance the separate terms in the loss, we set the disparity smoothness weight $\gamma = 0.001$ and the velocity loss weight $\lambda = 0.05$.

During adaptation, we utilize the same hyperparameters as listed above. Inspired by the findings of McCraith~\textit{et~al.}~\cite{mccraith2020monocular}, we freeze the weights of the encoders. Based on the ablation study in \refsec{ssec:ablation}, we set the number of update cycles $c=5$. To enhance the unsupervised guidance, we use the velocity readings to skip new incoming images if the driven distance is less than $\SI{0.2}{\meter}$.
We construct the training batch for the generalizer by concatenating the online data with a randomly sampled image triplet of each environment except for the current environment as this is already represented by the online data. Finally, we add the online data to the replay buffer.


{\parskip=3pt
\noindent\textbf{Datasets:}
To simulate scenes from a diverse set of environments, we employ our method on three relevant datasets, namely Cityscapes~\cite{cordts2016the}, Oxford RobotCar~\cite{maddern2017the}, and KITTI~\cite{geiger2012are}, posing the additional challenge of adapting to changing camera characteristics.
}


{\parskip=3pt
\noindent\textit{Cityscapes:}
The Cityscapes Dataset~\cite{cordts2016the} includes images and vehicle metadata recorded in 50 cities across Germany and bordering regions. Due to the unsupervised training scheme of our VO method, we can leverage the included 30-frame snippets to pre-train our networks despite the lack of ground truth poses.
}


{\parskip=3pt
\noindent\textit{Oxford RobotCar:}
The Oxford RobotCar Dataset~\cite{maddern2017the} focuses on repeated data recordings of a consistent route, captured over the period of one year in Oxford, UK. Besides RGB images, it also contains GNSS and IMU data, which we use for velocity supervision. To compute the trajectory error, we leverage the released RTK ground truth positions.
}


{\parskip=3pt
\noindent\textit{KITTI:}
The KITTI Dataset~\cite{geiger2012are} provides various sensor recordings taken in Karlsruhe, Germany. We utilize the training data from the odometry benchmark, which includes images and ground truth poses for multiple routes. We further leverage the corresponding IMU data from the released raw dataset to obtain the velocity of the vehicle.
}


\begin{table}[t]
\centering
\footnotesize
\caption{Path accuracy on the KITTI dataset}
\vspace{-0.15cm}
\label{tab:slam_metrics}
\begin{threeparttable}
    \setlength\tabcolsep{3pt}
    \begin{tabular}{ c | c c c  c c c  c c c  c c | c c }
    \toprule
        & \multicolumn{5}{c}{Online adaptation to KITTI} && \multicolumn{5}{c|}{Trained on KITTI seq. \{0, 1, 2, 8, 9\}} & \multicolumn{2}{c}{No training} \\
        \cmidrule(lr){2-6} \cmidrule(lr){8-12} \cmidrule(lr){13-14}
        KITTI & \multicolumn{2}{c}{CL-SLAM} && \multicolumn{2}{c}{CL-SLAM} && \multicolumn{2}{c}{DeepSLAM~\cite{li2021deepslam}} && \multicolumn{2}{c|}{VO+vel~\cite{godard2019digging, guizilini20203d}} & \multicolumn{2}{c}{ORB-SLAM} \\
        sequence & & && \multicolumn{2}{c}{(w/o loops)} && & && \multicolumn{2}{c}{(w/o loops)} \\
    \cmidrule(lr){2-3} \cmidrule(lr){5-6} \cmidrule(lr){8-9} \cmidrule(lr){11-12} \cmidrule(lr){13-14}
        & $t_{\mathit{err}}$ & $r_{\mathit{err}}$ && $t_{\mathit{err}}$ & $r_{\mathit{err}}$ && $t_{\mathit{err}}$ & $r_{\mathit{err}}$ && $t_{\mathit{err}}$ & $r_{\mathit{err}}$ & $t_{\mathit{err}}$ & $r_{\mathit{err}}$ \\
    \midrule
        4 & -- & -- && \textbf{4.37} & \textbf{0.51} && 5.22 & 2.27 && 10.72 & 1.69 & 0.62 & 0.11 \\
        5 & 4.30 & \textbf{1.01} && 4.41 & 1.33 && \textbf{4.04} & 1.40 && 34.55 & 11.88 & 2.51 & 0.25 \\
        6 & \textbf{2.53} & \textbf{0.63} && 3.07 & 0.73 && 5.99 & 1.54 && 15.20 & 5.62 & 7.80 & 0.35 \\
        7 & \textbf{2.10} & \textbf{0.83} && 3.74 & 1.91 && 4.88 & 2.14 && 12.77 & 6.80 & 1.53 & 0.35 \\
        10 & -- & -- && \textbf{2.22} & \textbf{0.34} && 10.77 & 4.45 && 55.27 & 9.50 & 2.96 & 0.52 \\
    \bottomrule
    \end{tabular}
    \vspace{1pt}
    Translation error $t_{\mathit{err}}$ in [\%] and rotation error $r_{\mathit{err}}$ in [\degree/100m].
    Sequences 4 and 10 do not contain loops.
    CL-SLAM is pre-trained on the Cityscapes dataset.
    The paths computed by ORB-SLAM use median scaling~\cite{zhou2017unsupervised} as they are not metric scale.
    The smallest errors among the learning-based methods are shown in bold.
\end{threeparttable}
\vspace*{-.5cm}
\end{table}

\subsection{Evaluation of Pose Accuracy of CL-SLAM}

Before analyzing how CL-SLAM addresses the task of continual SLAM, we compare its performance to existing SLAM framework. In particular, in \reftab{tab:slam_metrics} we report the translation and rotation errors on sequences 4, 5, 6, 7, and 10 of the KITTI Odometry dataset~\cite{geiger2012are} following Li~\textit{et~al.}~\cite{li2021deepslam}. Since the IMU data of sequence 3 has not been released, we omit this sequence.
We compare CL-SLAM to two learning-based and one feature-based approach. DeepSLAM~\cite{li2021deepslam} uses a similar unsupervised learning-based approach consisting of VO and graph optimization but does not perform online adaptation.
VO+vel refers to Monodepth2~\cite{godard2019digging} with velocity supervision~\cite{guizilini20203d}, i.e., it corresponds to the base VO estimator of CL-SLAM without adaptation and loop closure detection.
Both learning-based methods produce metric scale paths and are trained on the sequences 0, 1, 2, 8, and 9. Further, we report the results of monocular ORB-SLAM~\cite{murartal2015orbslam} after median scaling~\cite{zhou2017unsupervised}.

CL-SLAM outperforms DeepSLAM on the majority of sequences highlighting the advantage of online adaptation. Note that CL-SLAM was not trained on KITTI data but was only exposed to Cityscapes before deployment. To show the effect of global loop closure detection, we report the error on sequences 5 to 7 both with and without graph optimization enabled. Note that sequences 4 and 10 do not contain loops. Compared to ORB-SLAM, CL-SLAM suffers from a higher rotation error but can improve the translation error in sequences 6 and 10.
The overall results indicate that general SLAM methods would benefit from leveraging online information to enhance performance.


\subsection{Evaluation of Continual SLAM}

\begin{table}[t]
\centering
\footnotesize
\caption{Translation and rotation error for computing the AQ and RQ metrics}
\vspace{-0.15cm}
\label{tab:kitti_metrics}
\begin{threeparttable}
    \setlength\tabcolsep{3pt}
    \begin{tabular}{ c | c c | c c c  c c c  c c c  c c }
    \toprule
        Used & Previous & Current & \multicolumn{2}{c}{$\mathcal{B}_{\mathit{fixed}}$} && \multicolumn{2}{c}{$\mathcal{B}_{\mathit{expert}}$} && \multicolumn{2}{c}{$\mathcal{B}_{\mathit{general}}$} && \multicolumn{2}{c}{CL-SLAM} \\
    \cline{4-5} \cline{7-8} \cline{10-11} \cline{13-14}
        for & scenes & scene & $t_{\mathit{err}}$ & $r_{\mathit{err}}$ && $t_{\mathit{err}}$ & $r_{\mathit{err}}$ && $t_{\mathit{err}}$ & $r_{\mathit{err}}$ && $t_{\mathit{err}}$ & $r_{\mathit{err}}$ \\
    \midrule
        \multirow{4}{*}{AQ} & $c_t$ & $k_1$ & 130.74 & 26.35 && \textbf{2.50} & \textbf{0.37} && \underline{7.21} & \underline{1.26} && \textbf{2.50} & \textbf{0.37} \\
        & $c_t$ & $r_1$ & 170.76 & 13.37 && \textbf{28.94} & \underline{5.63} && \underline{29.05} & \textbf{5.49} && \textbf{28.94} & \underline{5.63} \\
        & $c_t \shortto r_1$ & $k_1$ & -- & -- && \underline{3.66} & \underline{0.73} && 14.14 & 1.79 && \textbf{3.24} & \textbf{0.54} \\
        & $c_t \shortto k_1$ & $r_1$ & -- & -- && \underline{32.56} & \underline{6.08} && 34.79 & 6.64 && \textbf{30.13} & \textbf{5.87} \\
    \midrule
        \multirow{4}{*}{RQ} & $c_t \shortto k_1 \shortto r_1$ & $k_2$ & 164.77 & 25.07 && 45.20 & 5.62 && \underline{8.48} & \underline{1.79} && \textbf{4.85} & \textbf{1.59} \\
        & $c_t \shortto k_1 \shortto r_1 \shortto k_2$ & $r_2$ & 200.14 & 28.94 && \textbf{15.91} & \underline{4.93} && \underline{16.02} & 4.98 && 20.50 & \textbf{4.77} \\
    \cmidrule{2-14}
        & $c_t \shortto k_1$ & $k_2$ & -- & -- && 15.82 & 2.50 && \underline{9.37} & \underline{2.21} && \textbf{7.48} & \textbf{1.63} \\
        & $c_t \shortto k_1 \shortto r_1$ & $r_2$ & -- & -- && \underline{14.89} & 4.62 && \textbf{12.24} & \textbf{4.38} && 16.41 & \underline{4.58} \\
    \bottomrule
    \end{tabular}
    \vspace{1pt}
    The \textit{previous scenes} denote the scenes that have been used for previous training of the algorithm, the \textit{current scene} denotes the evaluation scene to compute both errors $t_{\mathit{err}}$ in [\%] and $r_{\mathit{err}}$ in [\degree/100m].
    $c_t$ refers to the Cityscapes training set. $r_i$ and $k_i$ are sequences from KITTI and the Oxford RobotCar dataset.
    Bold and underlined values indicate the best and second best scores on each sequence.
\end{threeparttable}
\vspace*{-.5cm}
\end{table}

\textbf{Experimental Setup:}
In order to quantitatively evaluate the performance of our proposed approach, we compute both the adaptation quality (AQ) and the retention quality (RQ) by deploying CL-SLAM and the baseline methods on a fixed sequence of scenes. In particular, we use the official training split of the Cityscapes dataset to initialize the DepthNet and PoseNet, using the parameters detailed in \refsec{ssec:implementation_details}. The pre-training step is followed by a total of four scenes of the Oxford RobotCar dataset and the KITTI dataset.
\vspace*{-.2cm}
\begin{equation}
\begin{aligned}
    ( c_t \to k_1 \to r_1 \to k_2 \to r_2 ),
\end{aligned}
\label{eqn:eval_seq}
\end{equation}

\vspace*{-.2cm}\noindent where $c_t$ refers to the Cityscapes training set.

Following the setup of Li~\textit{et~al.}~\cite{li2020self}, we set $k_1$ and $k_2$ to be sequences~9 and 10 of the KITTI Odometry dataset. Note that we omit loop closure detection for this evaluation to prevent graph optimization from masking the effect of the respective adaptation technique. From the Oxford RobotCar dataset, we select the recording of August 12, 2015, at 15:04:18 GMT due to sunny weather and good GNSS signal reception. In detail, we set $r_1$ to be the scene between frames 750 and 4,750 taking every second frame to increase the driven distance between two consecutive frames. Analogously, we set $r_2$ to be the scene between frames 22,100 and 26,100. We use a scene length of 2,000 frames in order to be similar to the length of KITTI sequences: 1,584 frames for $k_1$ and 1,196 for $k_2$.


{\parskip=3pt
\noindent\textit{Baselines:}
We compare CL-SLAM to three baselines that are inspired by previous works towards online adaptation to a different environment compared to the environment used during training. As noted in \refsec{ssec:problem_setting}, continual SLAM differentiates from such a setting in the sense that it considers a sequence of different environments. First, $\mathcal{B}_{\mathit{expert}}$ imitates the strategy employed by Li~\textit{et~al.}~\cite{li2020self}, using a single set of network weights that is continuously updated based on the current data. This corresponds to only using the expert network in our architecture without resetting the weights. Second, $\mathcal{B}_{\mathit{general}}$ follows CoMoDA~\cite{kuznietsov2021comoda} leveraging a replay buffer built from previously seen environments. This method corresponds to only using the generalizer network. Finally, we compute the error without performing any adaptation, i.e., $\mathcal{B}_{\mathit{fixed}}$ utilizes network weights fixed after the pre-training stage.
To further illustrate forward and backward transfer and to close the gap to classical CL, we provide results on an additional baseline $\mathcal{B}_{\mathit{offline}}$ in the supplementary material. This baseline is initialized with the same network weights as CL-SLAM but does not perform online adaptation to avoid masking backward transfer. In reality, it resembles data collection followed by offline training after every new environment.
}


\begin{figure}[t]
\begin{minipage}{.4\textwidth}
    \footnotesize
    \centering
    \captionsetup{justification=centering}
    \captionof{table}{Comparison of the \\ Adaptation Quality (AQ)}
    \vspace{-0.15cm}
    \label{tab:adaptation_quality}
    \begin{threeparttable}
        \setlength\tabcolsep{5pt}
        \begin{tabular}{ l | c c }
            \toprule
                & $\uparrow$ AQ\textsubscript{trans} & $\uparrow$ AQ\textsubscript{rot} \\
             \midrule
                 $\mathcal{B}_{\mathit{fixed}}$ & 0.000 & 0.890 \\
                 $\mathcal{B}_{\mathit{expert}}$  & \underline{0.831} & \underline{0.982} \\
                 $\mathcal{B}_{\mathit{general}}$ & 0.787 & 0.979 \\
                 CL-SLAM & \textbf{0.848} & \textbf{0.983} \\
             \bottomrule
        \end{tabular}
        AQ\textsubscript{trans} refers to adaptation quality with respect to the translation error, AQ\textsubscript{rot} is based on the rotation error. Bold and underlined values denote the best and second best scores.
    \end{threeparttable}
\end{minipage}%
\hfill
\begin{minipage}{.575\textwidth}
    \centering
    \includegraphics[width=\linewidth]{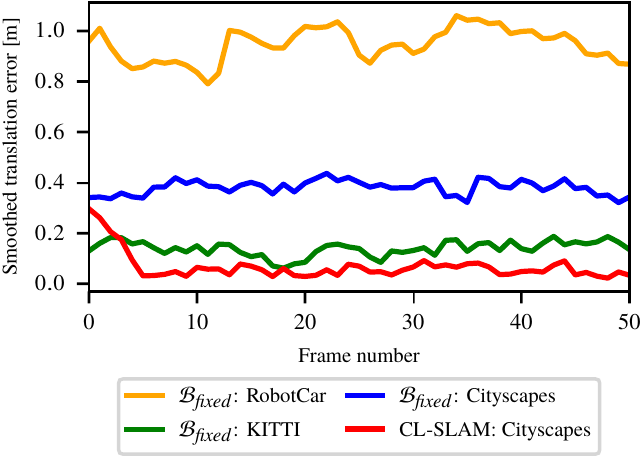}
    \captionof{figure}{The translation error on the initial frames of KITTI sequence 4. $\mathcal{B}_{\mathit{fixed}}$ is trained on the different environments indicating the domain gap between them. CL-SLAM overcomes this issue by performing online adaptation.}
    \label{fig:domain_gap}
\end{minipage}
\vspace{-.5cm}
\end{figure}

{\parskip=3pt
\noindent\textbf{Adapting to New Environments:}
In the initial part of the evaluation sequence (\refeqn{eqn:eval_seq}), the algorithm has to adapt to unseen environments. In accordance to the definition of the AQ in \refsec{ssec:metrics}, we construct four sequences listed in the upper four rows of \reftab{tab:kitti_metrics}.
Next, we deploy CL-SLAM and the baselines, initialized with the same set of model weights pre-trained on Cityscapes, on each of these sequences and compute the translation and rotation errors. Note that we do not apply median scaling since the PoseNet in our work produces metric estimates due to the velocity supervision term. Further note that for the first deployment after pre-training, $\mathcal{B}_{\mathit{expert}}$ corresponds to CL-SLAM. We observe that $\mathcal{B}_{\mathit{expert}}$ yields smaller errors than $\mathcal{B}_{\mathit{general}}$. This indicates the importance of online adaptation without diluting the updates with data from unrelated environments, if a high performance on the current deployment is the desideratum, and, thus, supports using the expert network in our approach. To compute the AQ score, after remapping using \refeqn{eqn:err_remapping} we sum the errors and divide by the number of sequences:}
\vspace*{-.2cm}
\begin{equation}
\begin{aligned}
    \text{AQ} &= \frac{1}{4} \left( M_{c_t \shortto k_1} + M_{c_t \shortto r_1} + M_{c_t \shortto r_1 \shortto k_1} + M_{c_t \shortto k_1 \shortto r_1} \right).
\end{aligned}
\end{equation}

\vspace*{-.2cm}
Comparing the AQ (see \reftab{tab:adaptation_quality}) for all experiments further endorses the previous findings in a single metric. Notably, continual adaptation is strictly necessary to obtain any meaningful trajectory.

Finally, we discuss the effect of consecutive deployments to different environments. In \reffig{fig:domain_gap}, we plot the translation error of the VO estimates on KITTI sequence 4 without online adaptation, separately trained on the considered datasets, and with adaptation, pre-trained on Cityscapes. As expected, without adaptation, the error is substantially higher if the system was trained on a different dataset showing the domain gap between the environments. By leveraging online adaptation, CL-SLAM reduces the initial error and yields even smaller errors than training on KITTI without further adaptation.
Having established the existence of a domain gap, we analyze how the deployment to the current environment effects the future deployment to another environment, resembling the concept of forward transfer (FTW) in continual learning (CL). In detail, \reftab{tab:kitti_metrics} reveals that the performances of all adaptation-based methods decrease when deploying them to an intermediate environment, e.g., $(c_t \to k_1)$ versus $(c_t \to r_1 \to k_1)$, where the effect is most pronounced for $\mathcal{B}_{\mathit{general}}$. In CL, such behavior is referred to as negative FWT.


\begin{figure}[t]
\centering
\includegraphics[width=\linewidth]{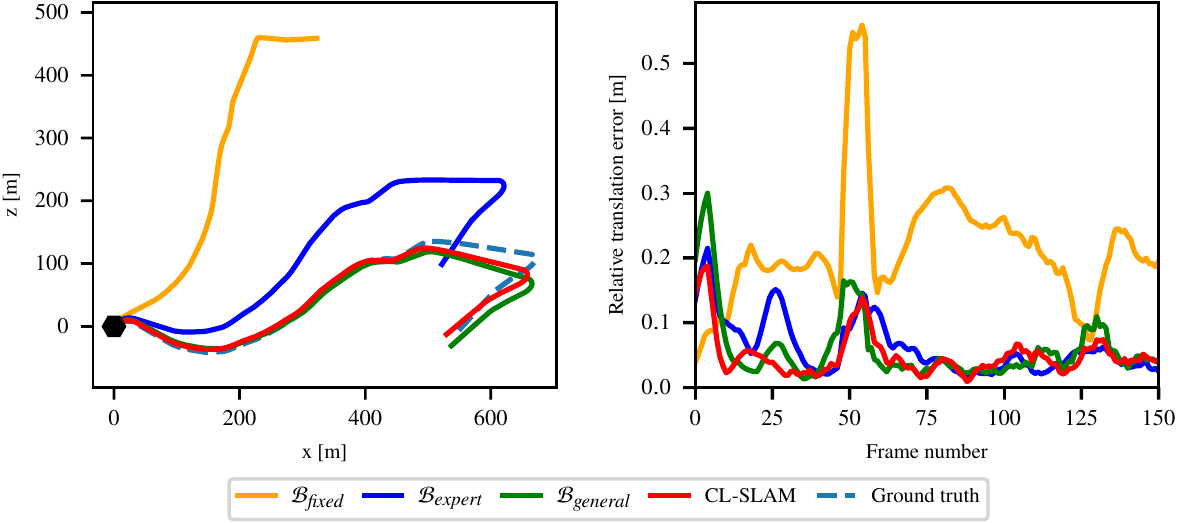}
\begin{minipage}[t]{.48\textwidth}
    \centering
    \captionof{figure}{Comparison of the trajectory on $k_2$ after previous deployment on $k_1$ and $r_1$ predicted by CL-SLAM and the baseline methods. The hexagon indicates the starting point.}
    \label{fig:retention_trajectory}
\end{minipage}%
\hfill
\begin{minipage}[t]{.48\textwidth}
    \centering
    \captionof{figure}{Relative translation error of the first 150 frames along $k_2$. Compared to $\mathcal{B}_{\mathit{expert}}$, CL-SLAM reduces the error more quickly due to initialization with the weights of its generalizer network.}
    \label{fig:retention_error}
\end{minipage}
\vspace*{-.5cm}
\end{figure}

{\parskip=3pt
\noindent\textbf{Remembering Previous Environments:}
In the subsequent phase of the evaluation sequence (\refeqn{eqn:eval_seq}), the algorithm is redeployed in a new scene taken from a previously encountered environment. In accordance to the definition of the RQ in \refsec{ssec:metrics}, we construct four sequences listed in the lower four rows of \reftab{tab:kitti_metrics}. Note that the first two sequences are part of the original evaluation sequence (\refeqn{eqn:eval_seq}) and the other two sequences are used as a reference to measure the effect of mixed environments.
}

Following the same procedure as in the previous section, we compute the translation and rotation errors. The resulting scores (see \reftab{tab:kitti_metrics}) demonstrate the benefit of employing a replay buffer to leverage previously learned knowledge, $\mathcal{B}_{\mathit{general}}$ yields smaller errors than $\mathcal{B}_{\mathit{expert}}$ on the majority of sequences.
To compute the RQ, we follow \refeqn{eqn:rq}:
\vspace*{-.2cm}
\begin{equation}
\begin{aligned}
    \text{RQ} &= \frac{1}{2} \bigl[ \left( M_{c_t \shortto k_1 \shortto r_1 \shortto k_2} - M_{c_t \shortto k_1 \shortto k_2} \right) + \left( M_{c_t \shortto k_1 \shortto r_1 \shortto k_2 \shortto r_2} - M_{c_t \shortto k_1 \shortto r_1 \shortto r_2} \right) \bigr].
\end{aligned}
\end{equation}

\vspace*{-.2cm}
Comparing the RQ scores in \reftab{tab:retention_quality} clearly shows that the drop in performance when mixing environments is less pronounced for $\mathcal{B}_{\mathit{general}}$. Our proposed CL-SLAM leverages this advantage due to its generalizer, while the expert still focuses on the current scene, achieving the highest RQ across the board.

To bridge the gap to classical CL, we also qualitatively compare the consecutive deployment to scenes from the same environment with introducing an intermediate scene from another environment, e.g., $(c_t \to k_1 \to k_2)$ versus $(c_t \to k_1 \to r_1 \to k_2)$. In CL, an increase/decrease in performance is known as positive/negative backward transfer (BWT). Whereas we observe positive BWT for $\mathcal{B}_{\mathit{general}}$ and CL-SLAM on the KITTI dataset, the sequence with final deployment on RobotCar suffers from negative BWT. A possible explanation for this inconsistent behavior is structural differences between the sequences of the same dataset inducing small domain gaps within a dataset that require a potentially more fine-grained scene classification. However, by always performing online adaptation independent of previous deployments, CL-SLAM circumvents such issues.

In \reffig{fig:retention_trajectory}, we visualize the generated trajectories in $k_2$ given previous deployment in $k_1$ and $r_1$ from our method and the evaluated baselines. Although $\mathcal{B}_{\mathit{expert}}$ can reproduce the general shape of the trajectory, it requires a warm-up time causing an initial drift, visible up to frame 40 in \reffig{fig:retention_error}. On the other hand, $\mathcal{B}_{\mathit{general}}$ can leverage the experience from $k_1$ due to the rehearsal of the KITTI data from its replay buffer during the previous deployment in $r_1$. By following this idea, our proposed CL-SLAM combines the advantages of both baseline strategies.

\begin{figure}[b]
\begin{minipage}{.45\textwidth}
    \footnotesize
    \centering
    \captionsetup{justification=centering}
    \captionof{table}{Comparison of the \\ Retention Quality (RQ)}
    \vspace{-0.15cm}
    \label{tab:retention_quality}
    \begin{threeparttable}
        \setlength\tabcolsep{2pt}
        \begin{tabular}{ l | c c }
            \toprule
                & $\uparrow$ RQ\textsubscript{trans} $\times 10^{-3}$ & $\uparrow$ RQ\textsubscript{rot} $\times 10^{-3}$ \\
            \midrule
                $\mathcal{B}_{\mathit{fixed}}$   & --                & --                \\
                $\mathcal{B}_{\mathit{expert}}$  & -152.0            & -9.5              \\
                $\mathcal{B}_{\mathit{general}}$ & \underline{-14.4} & \underline{-0.5}  \\
                CL-SLAM                          & \textbf{-7.3}     & \textbf{-0.4}     \\
            \bottomrule
        \end{tabular}
        \vspace{1pt}
        RQ\textsubscript{trans} refers to the retention quality with respect to the translation error, RQ\textsubscript{rot} is based on the rotation error. $\mathcal{B}_{\mathit{fixed}}$ does not perform adaptation, hence computing the RQ is meaningless. Bold and underlined values denote the best and second best scores.
    \end{threeparttable}
\end{minipage}%
\hfill
\begin{minipage}{.52\textwidth}
    \footnotesize
    \centering
    \captionsetup{justification=centering}
    \captionof{table}{Ablation study on the number \\ of adaptation cycles}
    \vspace{-0.15cm}
    \label{tab:ablation_iters}
    \begin{threeparttable}
        \setlength\tabcolsep{2pt}
        \begin{tabular}{ c  c | c c c  c c}
            \toprule
             & & \multicolumn{2}{c} {$c_t \to k_1$} && \multicolumn{2}{c}{$c_t \to k_2$} \\
            \cmidrule(lr){3-4} \cmidrule(lr){6-7}
            Updates & Relative FPS & $t_{\mathit{err}}$ & $r_{\mathit{err}}$ && $t_{\mathit{err}}$ & $r_{\mathit{err}}$ \\
            \midrule
            1 & 1.00 & 34.37 & 6.70 && 86.36 & 11.71 \\
            2 & 0.56 & 31.37 & 5.83 && 39.72 &  7.16 \\
            3 & 0.40 & 24.21 & 4.21 && \textbf{11.15} &  4.63 \\
            4 & 0.30 &  3.24 & 0.54 && 13.51 &  2.03 \\
            5 & 0.24 & \textbf{2.50} & \textbf{0.37} && \underline{11.18} & \underline{1.74} \\
            6 & 0.20 & \underline{2.84} & \underline{0.40} && 12.97 & \textbf{1.51} \\
            \bottomrule
        \end{tabular}
        \vspace{1pt}
        Translation error $t_{\mathit{err}}$ in [\%] and rotation error $r_{\mathit{err}}$ in [\degree/100m] for varying number of weight updates $c$ performed during online adaptation. We use $c=5$ in CL-SLAM. Bold and underlined values denote the best and second best scores.
    \end{threeparttable}
\end{minipage}
\end{figure}


\label{ssec:ablation}
{\parskip=3pt
\noindent\textbf{Number of Update Cycles:}
We perform a brief ablation study on the number of update cycles performed during online adaptation, i.e., how often steps (2) to (4) are repeated for a given batch of data (see \refsec{ssec:online_adaptation}). For this, we deploy CL-SLAM to both KITTI sequences $k_1$ and $k_2$ and compute the translation and rotation error. As shown in \reftab{tab:ablation_iters}, using five update cycles yields the most accurate trajectory while resulting in a 75\% reduction in speed compared to a single cycle. However, please note that in this work, we do not focus on adaptation speed but on showing the efficacy of the proposed dual-network approach to balance the common continual learning trade-off between quick adaptation and memory retention.
}

\section{Conclusion}

In this paper, we introduced the task of continual SLAM, which requires the SLAM algorithm to continuously adapt to new environments while retaining the knowledge learned in previously visited environments. To evaluate the capability of a given model to meet these opposing objectives, we defined two new metrics based on the commonly used translation and rotation errors, namely the adaptation quality and the retention quality. As a potential solution, we propose CL-SLAM, a deep learning-based visual SLAM approach that predicts metric scale trajectories from monocular videos and detects global loop closures. To balance short-term adaptation and long-term memory retention, CL-SLAM is designed as a dual-network architecture comprising an expert and a generalizer, which leverages experience replay. Through extensive experimental evaluations, we demonstrated the efficacy of our method compared to baselines using previously proposed continual learning strategies for online adaptation of visual odometry. Future work will focus on transferring the proposed design scheme to more advanced visual odometry methods, e.g., using point matching via optical flow. We further plan to address the currently infinite replay buffer to mitigate the scaling problem, e.g., by storing more abstract representations or keeping only the most representative images.


\begin{footnotesize}
    \bibliographystyle{spmpsci}
    \bibliography{references.bib}
\end{footnotesize}


\appendix
\clearpage
\setcounter{section}{0}
\renewcommand\thesection{\Alph{section}}
\section{Supplementary Material}


\subsection{Technical Approach: Visual Odometry}

As elaborated in the main paper, we generate VO estimates following the commonly used approach of using a trainable pose network~\citeSM{besic2022dynamic_, godard2019digging_, guizilini20203d_, li2021deepslam_} for self-supervised depth estimation with a stream of monocular images. In this section, we first review the basic idea behind this approach and then describe the losses that we employ in more detail.

The core intuition is that given a source image $\mathbf{I_s}$ and the camera motion $\mathbf{O_{s \shortto t}}$, it is possible to generate a reconstructed view $\mathbf{\hat I_{s \shortto t}}$ for a target image $\mathbf{I_t}$ using image warping. In detail, a 2D pixel $\mathbf{p_t}$ can be projected to the 3D point $\mathbf{P_t}$ using the camera matrix $\mathbf{K}$ and depth information $\mathbf{d_t}$ at this pixel. In monocular depth estimation, $\mathbf{d_t}$ is predicted by a neural network. Next, the camera motion $\mathbf{O_{s \shortto t}}$ is used to transform $\mathbf{P_t}$ to $\mathbf{P_s}$, which can be projected onto the plane of image $\mathbf{I_s}$ yielding the 2D pixel $\mathbf{\hat p_s}$:
\begin{equation}
    \mathbf{\hat p_s} \sim \mathbf{K O_{t \shortto s} \underbrace{\mathbf{d_t K^{-1} p_t}}_{\mathbf{P_t}}}.
\end{equation}

Repeating this procedure for every pixel in $\mathbf{I_t}$, we obtain a mapping $\mathbf{p_t} \mapsto \mathbf{\hat p_s}$ to reproject image coordinates:
\begin{equation}
    \mathbf{\hat I_{s \shortto t}(p_t)} = \mathbf{I_s(\hat p_s)}.
\end{equation}

Finally, using bilinear interpolation over these coordinates, the reconstructed view $\mathbf{\hat I_{s \shortto t}}$ can be generated and compared to the target image $\mathbf{I_t}$ to compute a loss value.

Our total loss is composed of the photometric reprojection loss $\mathcal{L}_{pr}$, the image smoothness loss $\mathcal{L}_{sm}$, and the velocity supervision loss $\mathcal{L}_{vel}$:
\begin{equation}
    \mathcal{L} = \mathcal{L}_{pr} + \gamma \mathcal{L}_{sm} + \lambda \mathcal{L}_{vel}.
\end{equation}


{\parskip=5pt
\noindent\textit{Photometric Consistency:}
To minimize the photometric error between the true target image and the reconstructed view, we compute the structural dissimilarity $\mathcal{L}_{sim}$~\citeSM{godard2017unsupervised_}:}
\begin{equation}
    \mathcal{L}_{sim}(\mathbf{I}, \mathbf{\hat I}) = \alpha \frac{1 - SSIM(\mathbf{I}, \mathbf{\hat I})}{2} + (1 - \alpha) ||\mathbf{I} - \mathbf{\hat I}||_1,
\end{equation}
where $SSIM$ denotes the structure similarity image matching index~\citeSM{wang2004image_}. To mitigate the effect of objects that are present in the target image $\mathbf{I_t}$ but not in the source images, Godard \textit{et al.}~\citeSM{godard2019digging_} proposed to take the pixel-wise minimum over $\mathcal{L}_{sim}(\mathbf{I_t}, \mathbf{\hat I_{s \shortto t}})$ for all source images:
\begin{equation}
    \mathcal{L}_p = \min_s \mathcal{L}_{sim}(\mathbf{I_t}, \mathbf{\hat I_{s \shortto t}}).
\end{equation}

To further suppress the signal from static scenes or objects moving at a similar speed as the ego-robot, the same authors introduced the concept of auto-masking. The idea is to compute the loss only on those pixels, where the photometric error $\mathcal{L}_{sim}(\mathbf{I_t}, \mathbf{\hat I_{s \shortto t}})$ of the reconstructed image is smaller than the error $\mathcal{L}_{sim}(\mathbf{I_t}, \mathbf{\hat I_{s}})$ of the original source image:
\begin{equation}
    \mu_{mask} = \left[ \min_s \mathcal{L}_{sim}(\mathbf{I_t}, \mathbf{\hat I_{s \shortto t}}) < \min_s \mathcal{L}_{sim}(\mathbf{I_t}, \mathbf{I_{s}}) \right],
\end{equation}
where $\mu_{mask}$ has the same width and height as $\mathbf{I_t}$.

The total photometric reprojection loss $\mathcal{L}_{pr}$ is defined as:
\begin{equation}
    \mathcal{L}_{pr} = \mu_{mask} \cdot \mathcal{L}_p.
\end{equation}


{\parskip=5pt
\noindent\textit{Image Smoothness:}
To regularize the depth prediction in image regions with less texture, we use an edge-aware smoothness term~\citeSM{godard2017unsupervised_} computed for the predicted depth map $\mathbf{D_t}$. It encourages the DepthNet to generate continuous depth values in continuous image areas.}
\begin{equation}
    \mathcal{L}_{sm} = |\partial_x \mathbf{S_t^*}| e^{-|\partial_x \mathbf{I_t}|} + |\partial_y \mathbf{S_t^*}| e^{-|\partial_y \mathbf{I_t}|},
\end{equation}
where $\partial_i$ indicates the partial derivative with respect to axis $i$ and $\mathbf{S_t^*} = \mathbf{S_t} / \mathbf{\bar S_t}$ denotes the inverse depth (disparity) $\mathbf{S_t} = \mathbf{D_t}^{-1}$ normalized with its mean.


{\parskip=5pt
\noindent\textit{Velocity Supervision:}
To enforce metric scaling of the predicted odometry, we leverage the speed or velocity measurements from the robot. While such a rough measurement can be obtained inexpensively, e.g., by wheel odometry, it teaches the network to predict scale-aware depth and pose estimates. The velocity supervision term $\mathcal{L}_{vel}$~\citeSM{guizilini20203d_} imposes a loss between the magnitude of the predicted translation $T_{t \shortto s}$ and the distance traveled by the robot based on the velocity reading $v_{t \shortto s}$:}
\begin{equation}
    \mathcal{L}_{vel} = \sum_s \Bigm| ||T_{t \shortto s}||_2 - |v_{t \shortto s}| \Delta \tau_{t \shortto s} \Bigm|,
\end{equation}
where $\Delta \tau_{t \shortto s}$ denotes the time between images $\mathbf{I_s}$ and $\mathbf{I_t}$.


\subsection{Additional Experimental Evaluation}

In \reftab{tab:sup_kitti_metrics}, we provide the results of an additional baseline that we call $\mathcal{B}_{\mathit{offline}}$. It does not leverage online adaptation but is initialized with the same network parameters as CL-SLAM. Note that in a practical setting, to some extent this corresponds to data collection followed by offline training with a replay buffer for every new environment. Although such a setup does not completely align with the core idea of continual SLAM, $\mathcal{B}_{\mathit{offline}}$ aims to close the gap to classical continual learning by not performing re-adaptation to previously seen environments, i.e., conducting a previously learned task, to avoid masking backward transfer.

Analogously to the adaptation-based methods, the performance of $\mathcal{B}_{\mathit{offline}}$ on $k_1$ degrades with an intermediate deployment to $r_1$, which is referred to as positive forward transfer (FWT) in classical continual learning (CL). Unlike the other methods, $\mathcal{B}_{\mathit{offline}}$ yields smaller errors on $r_1$ if it was previously deployed to $k_1$, known as positive FWT. A possible explanation for this inconsistent behavior is structural differences between the sequences of the same dataset inducing small domain gaps within a dataset that require a potentially more fine-grained scene classification.

\begin{table}
\centering
\footnotesize
\caption{Translation and rotation error for computing the AQ and RQ metrics}
\vspace{-0.15cm}
\label{tab:sup_kitti_metrics}
\begin{threeparttable}
    \setlength\tabcolsep{5pt}
    \begin{tabular}{ c c | c c c  c c }
    \toprule
       Previous & Current & \multicolumn{2}{c}{$\mathcal{B}_{\mathit{offline}}$} && \multicolumn{2}{c}{CL-SLAM} \\
    \cline{3-4} \cline{6-7}
        scenes & scene & $t_{\mathit{err}}$ & $r_{\mathit{err}}$ && $t_{\mathit{err}}$ & $r_{\mathit{err}}$ \\
    \midrule
        $c_t$ & $k_1$ & 130.74 & 26.35 && 2.50 & 0.37 \\
        $c_t$ & $r_1$ & 170.76 & 13.37 && 28.94 & 5.63 \\
        $c_t \shortto r_1$ & $k_1$ & 182.62 & 38.38 && 3.24 & 0.54 \\
        $c_t \shortto k_1$ & $r_1$ & 52.23 & 8.49 && 30.13 & 5.87 \\
    \midrule
        $c_t \shortto k_1 \shortto r_1$ & $k_2$ & 23.77 & 4.76 && 4.85 & 1.59 \\
        $c_t \shortto k_1 \shortto r_1 \shortto k_2$ & $r_2$ & 105.54 & 25.64 && 20.50 & 4.77 \\
    \midrule
        $c_t \shortto k_1$ & $k_2$ & 69.48 & 7.45 && 7.48 & 1.63 \\
        $c_t \shortto k_1 \shortto r_1$ & $r_2$ & 153.77 & 35.21 && 16.41 & 4.58 \\
    \bottomrule
    \end{tabular}
    \vspace{1pt}
    The \textit{previous scenes} denote the scenes that have been used for previous training of the algorithm, the \textit{current scene} denotes the evaluation scene to compute both errors $t_{\mathit{err}}$ in [\%] and $r_{\mathit{err}}$ in [\degree/100m].
\end{threeparttable}
\end{table}

\begin{figure}
    \centering
    \includegraphics[width=.48\textwidth]{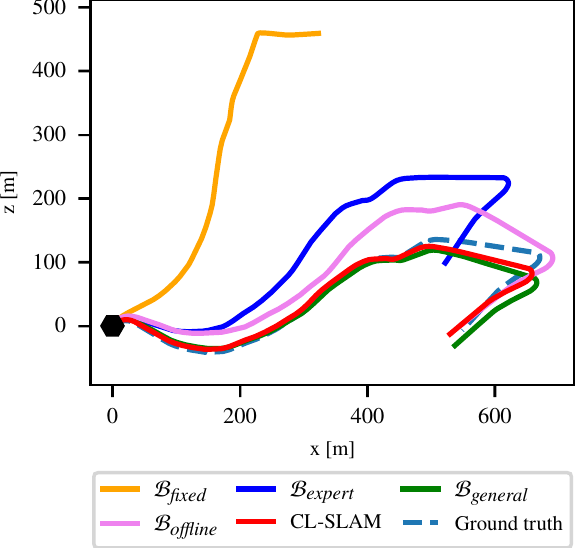}
    \caption{A modified version of \reffig{fig:retention_trajectory} including the additional baseline $\mathcal{B}_{\mathit{offline}}$. Comparison of the trajectory on $k_2$ after previous deployment on $k_1$ and $r_1$ predicted by CL-SLAM and the baseline methods.}
    \label{fig:retention_trajectory_supp}
    \vspace*{-.2cm}
\end{figure}

\begin{footnotesize}
    \bibliographystyleSM{spmpsci}
    \bibliographySM{references_supplement.bib}
\end{footnotesize}


\end{document}